\documentclass[10pt,twocolumn,letterpaper]{article}

\usepackage{iccv}
\usepackage{times}
\usepackage{epsfig}
\usepackage{graphicx}
\usepackage{amsmath}
\usepackage{amssymb}
\usepackage[square,numbers,sort&compress]{natbib}

\usepackage{subfigure}
\usepackage{upgreek}
\usepackage{multirow}
\usepackage{color}
\usepackage{bm}

\usepackage{arydshln}
\usepackage{latexsym}

\usepackage{amsthm}
\newtheorem{theorem}{Theorem}

\usepackage{authblk}
\usepackage{xpatch}
\usepackage{lipsum}
\let\OLDthebibliography\thebibliography
\renewcommand\thebibliography[1]{
  \OLDthebibliography{#1}
  \setlength{\parskip}{0pt}
  \setlength{\itemsep}{0.2pt plus 0.05ex}
}

\usepackage[pagebackref=true,breaklinks=true,letterpaper=true,colorlinks,bookmarks=false]{hyperref}
\renewcommand{\huge}{\fontsize{8.31pt}{\baselineskip}\selectfont}

\usepackage[breaklinks=true,bookmarks=false]{hyperref}

\iccvfinalcopy 

\def\iccvPaperID{****} 

\begin{document} 

\title{Multi-channel Weighted Nuclear Norm Minimization 
\\
for Real Color Image Denoising}
\vspace{-4mm}
\xpatchcmd{\author}{\relax#1\relax}{\relax\detokenize{#1}\relax}{}{}
\author[1]{Jun Xu}
\author[\empty]{Lei Zhang\textsuperscript{1,}\thanks{This work is supported by HK RGC GRF grant (PolyU 5313/13E) and China NSFC grant (no. 61672446).}}
\author[1]{David Zhang}
\author[2]{Xiangchu Feng}
\affil[1]{Dept. of Computing, The Hong Kong Polytechnic University, Hong Kong, China}
\affil[2]{School of Mathematics and Statistics, Xidian University, Xi'an, China 
\authorcr{\tt \small{\{csjunxu,cslzhang,csdzhang\}@comp.polyu.edu.hk,xcfeng@mail.xidian.edu.cn}}
}
\vspace{-4mm}

\maketitle
\vspace{-4mm}

\begin{abstract}
Most of the existing denoising algorithms are developed for grayscale images.\ It is not trivial to extend them for color image denoising since the noise statistics in R, G, and B channels can be very different for real noisy images.\ In this paper, we propose a multi-channel (MC) optimization model for real color image denoising under the weighted nuclear norm minimization (WNNM) framework.\ We concatenate the RGB patches to make use of the channel redundancy, and introduce a weight matrix to balance the data fidelity of the three channels in consideration of their different noise statistics.\ The proposed MC-WNNM model does not have an analytical solution.\ We reformulate it into a linear equality-constrained problem and solve it via alternating direction method of multipliers.\ Each alternative updating step has a closed-form solution and the convergence can be guaranteed.\ Experiments on both synthetic and real noisy image datasets demonstrate the superiority of the proposed MC-WNNM over state-of-the-art denoising methods. 
\end{abstract}

\vspace{-3mm}
\section{Introduction}
\vspace{-2mm}

Image denoising is a classical yet fundamental problem for image quality enhancement in computer vision and photography systems.\ Most of existing denoising algorithms are designed for grayscale images, aiming to recover the clean image $\mathbf{x}$ from its noisy observation $\mathbf{y}=\mathbf{x}+\mathbf{n}$, where $\mathbf{n}$ is generally assumed to be additive white Gaussian noise (AWGN).\ State-of-the-art image denoising methods include sparse representation \cite{bm3d}, dictionary learning \cite{ksvd}, low-rank approximation \cite{wnnm}, non-local self-similarity (NSS) \cite{nlm} based methods, and the combination of those techniques \cite{ksvd,bm3d,lssc,epll,ncsr,pgpd,wnnm}.\ Recently, some discriminative denoising methods have also been developed by learning discriminative priors from pairs of clean and noisy images \cite{mlp,csf,chen2015learning,dncnn}. 

When the input is a noisy RGB color image, there are mainly three strategies for color image denoising.\ (1) The first strategy is to apply the grayscale image denoising algorithm to each channel.\ However, such a straightforward solution will not exploit the spectral correlation among RGB channels, and the denoising performance may not be very satisfying.\ (2) The second strategy is to transform the RGB image into a less correlated color space, such as YCbCr, and perform denoising in each channel of the transformed space \cite{foe,cbm3d}.\ One representative work along this line is the CBM3D algorithm \cite{cbm3d}.\ However, the color transform will complicate the noise distribution, and the correlation among color channels is not fully exploited.\ (3) The third strategy is to perform joint denoising on the RGB channels simultaneously for better use of the spectral correlation.\ For example, the patches from RGB channels are concatenated as a long vector for processing \cite{mairal2008sparse,Zhu_2016_CVPR}. 

Though joint denoising of RGB channels is a more promising way for color image denoising, it is not a trivial extension from single channel (grayscale image) to multiple channels (color image).\ The noise in standard RGB (sRGB) space can be approximately modeled as AWGN, but it has different variances for different channels \cite{Liu2008,Leungtip,crosschannel2016} due to the sensor characteristics and on-board processing steps in digital camera pipelines \cite{crosschannel2016,karaimer_brown_ECCV_2016}.\ This makes the real color image denoising problem much more complex.\ If the three channels are treated equally in the joint denoising process, false colors or artifacts can be generated \cite{mairal2008sparse}.\ How to account for the different noise characteristics in color channels, and how to effectively exploit the within and cross channel correlation are the key issues for designing a good color image denoising method.

This paper presents a new color image denoising algorithm.\ Considering that the weighted nuclear norm minimization (WNNM) method \cite{wnnm,wnnmijcv}, which exploits the image NSS property via low rank regularization, has achieved excellent denoising performance on grayscale images, we propose to extend WNNM to real color image denoising.\ More specifically, we propose a multi-channel WNNM (MC-WNNM) model, which concatenates the patches from RGB channels for rank minimization but introduces a weight matrix to adjust the contributions of the three channels based on their noise levels.\ The proposed MC-WNNM model no longer has a closed-form solution as in the original WNNM model \cite{wnnmijcv}.\ We reformulate it into a linear equality-constrained problem with two variables, and solve the relaxed problem under the alternating direction method of multipliers \cite{admm} framework.\ Each variable can be updated with closed-form solutions, and the convergence analysis is given to guarantee a rational termination of the proposed algorithm.

\vspace{-1mm}
\section{Related Work}

\subsection{Weighted Nuclear Norm Minimization}
\vspace{-1mm}

As a generalization to the nuclear norm minimization (NNM) model \cite{cai2010singular}, the weighted nuclear norm minimization (WNNM) model \cite{wnnm,wnnmijcv} is described as 
\vspace{-3mm}
\begin{equation}
\label{e1}
\vspace{-3mm}
\min_{\mathbf{X}}\|\mathbf{Y}-\mathbf{X}\|_{F}^{2}
+
\|\mathbf{X}\|_{\bm{w},*},
\end{equation}
where $\|\mathbf{X}\|_{\bm{w},*}=\sum_{i}w_{i}\sigma_{i}(\mathbf{X})$ is the weighted nuclear norm of matrix $\mathbf{X}$, $\bm{w}=[w_{1},...,w_{n}]^{\top}$ ($w_{i}\ge 0$) is the weight vector, and $\sigma_{i}(\mathbf{X})$ is the $i$th singular value of $\mathbf{X}$.\ According to the Corollary 1 of \cite{wnnmijcv}, if the weights are non-decreasing, the problem (\ref{e1}) has a closed-form solution:
\vspace{-3mm}  
\begin{equation}
\label{e2}
\vspace{-3mm}
\mathbf{\hat{X}}
=
\mathbf{U}
\mathcal{S}_{\bm{w}/2}
(\mathbf{\Sigma})
\mathbf{V}^{\top},
\end{equation}
where $\mathbf{Y}=\mathbf{U}\mathbf{\Sigma}\mathbf{V}^{\top}$ is the singular value decomposition (SVD) \cite{eckart1936approximation} of $\mathbf{Y}$ and 
$\mathcal{S}_{\bm{w}/2}(\bullet)$ is the generalized soft-thresholding operator with weight vector $\bm{w}$:
\vspace{-3mm}
\begin{equation}
\label{e3}
\vspace{-3mm}
\mathcal{S}_{\bm{w}/2}
(\mathbf{\Sigma}_{ii})
=
\max(\mathbf{\Sigma}_{ii}-w_{i}/2, 0).
\end{equation}

WNNM has demonstrated highly competitive denoising performance on grayscale images.\ However, if we directly extend it to color image denoising by concatenating the patches from RGB channels, denoising artifacts may happen (please refer to Fig.\ \ref{f1} and the section of experimental results).\ In this paper, we propose a multi-channel WNNM (MC-WNNM) model for color image denoising, which preserves the power of WNNM and is able to address the noise differences among different channels.

\vspace{-1mm}
\subsection{Real Color Image Denoising}
\vspace{-1mm}

During the last decade, a few methods have been proposed for real color image denoising.\ Among them, the CBM3D method \cite{cbm3d} is a representative one, which first transforms the RGB image into a luminance-chrominance space (e.g., YCbCr) and then applies the benchmark BM3D method \cite{bm3d} to each channel separately.\ The non-local similar patches are grouped by the luminance channel.\ In \cite{Liu2008}, Liu et al.\ proposed the ``Noise Level Function'' to estimate and remove the noise for each channel in natural images.\ However, processing each channel separately would often achieve inferior performance to processing the color channels jointly \cite{mairal2008sparse}.\ Therefore, the methods \cite{noiseclinic,ncwebsite,Zhu_2016_CVPR} perform real color image denoising by concatenating the patches of RGB channels into a long vector.\ However, the concatenation treats each channel equally and ignores the different noise statistics among these channels.\ The method in \cite{crosschannel2016} models the cross-channel noise in real noisy images as multivariate Gaussian and the noise is removed by the Bayesian non-local means filter \cite{kervrann2007bayesian}.\ The commercial software Neat Image \cite{neatimage} estimates the noise parameters from a flat region of the given noisy image and filters the noise accordingly.\ The methods in \cite{crosschannel2016,neatimage} ignore the non-local self-similarity of natural images \cite{bm3d,wnnm}. 

In this paper, we present an effective multi-channel image denoising algorithm, which utilizes the strong low-rank prior of image non-local similar patches, and introduces a weight matrix to balance the multi-channels based on their different noise levels.

\vspace{-1mm}
\section{\parbox[t]{10cm}{The Proposed Color Image Denoising Algo-\\rithm}}

\subsection{\parbox[t]{10cm}{The Multi-channel Weighted Nuclear Norm \\Minimization Model}}
\vspace{-0mm}

\begin{figure*}
\vspace{-5mm}
\centering
\subfigure{
\begin{minipage}[t]{0.25\textwidth}
\centering
\raisebox{-0.15cm}{\includegraphics[width=1\textwidth]{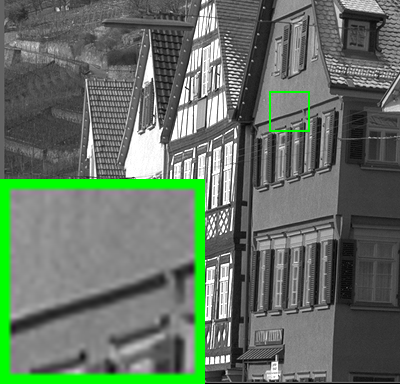}}
{\footnotesize (a) Clean Red Channel}
\end{minipage}
\begin{minipage}[t]{0.25\textwidth}
\centering
\raisebox{-0.15cm}{\includegraphics[width=1\textwidth]{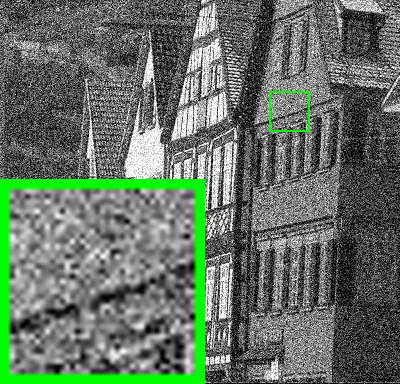}}
{\footnotesize (b) Noisy Red Channel}
\end{minipage}
\begin{minipage}[t]{0.25\textwidth}
\centering
\raisebox{-0.15cm}{\includegraphics[width=1\textwidth]{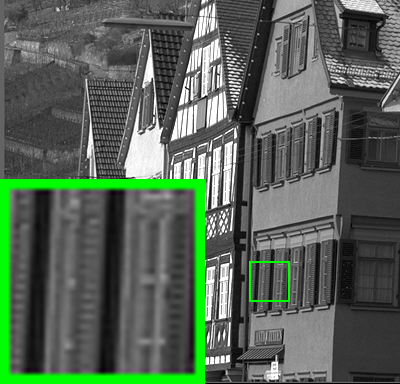}}
{\footnotesize (c) Clean Green Channel}
\end{minipage}
\begin{minipage}[t]{0.25\textwidth}
\centering
\raisebox{-0.15cm}{\includegraphics[width=1\textwidth]{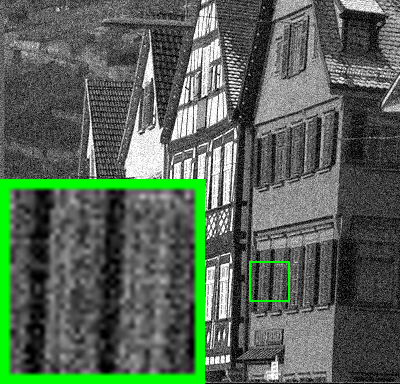}}
{\footnotesize (d) Noisy Green Channel}
\end{minipage}
}\vspace{-3mm}
\subfigure{
\begin{minipage}[t]{0.25\textwidth}
\centering
\raisebox{-0.15cm}{\includegraphics[width=1\textwidth]{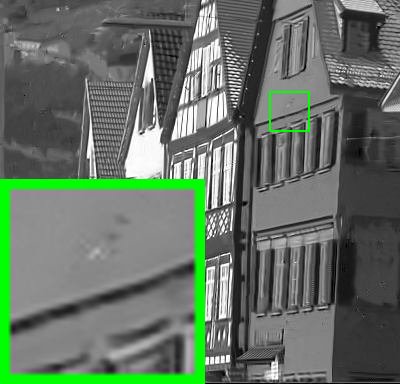}}
{\footnotesize (e) Denoised Red by WNNM }
\end{minipage}
\begin{minipage}[t]{0.25\textwidth}
\centering
\raisebox{-0.15cm}{\includegraphics[width=1\textwidth]{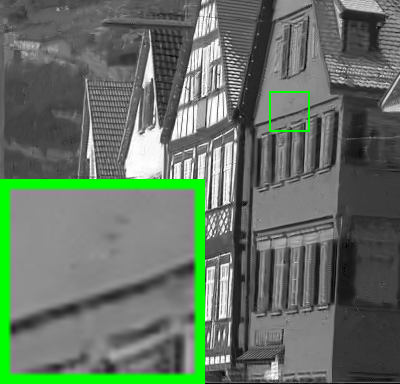}}
{\footnotesize (f) Denoised Red by MC-WNNM }
\end{minipage}
\begin{minipage}[t]{0.25\textwidth}
\centering
\raisebox{-0.15cm}{\includegraphics[width=1\textwidth]{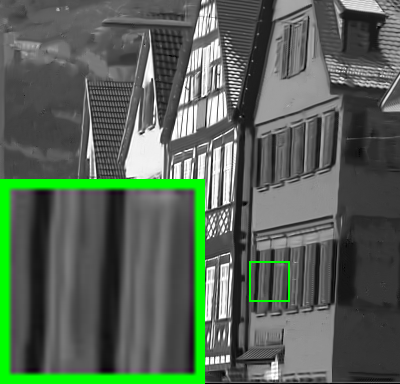}}
{\footnotesize (g) Denoised Green by WNNM }
\end{minipage}
\begin{minipage}[t]{0.25\textwidth}
\centering
\raisebox{-0.15cm}{\includegraphics[width=1\textwidth]{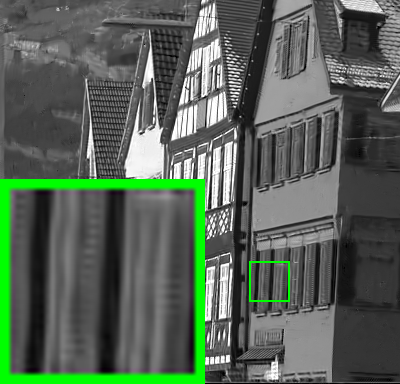}}
{\footnotesize (h) Denoised Green by MC-WNNM }
\end{minipage}
}\vspace{-1mm}
\caption{The Red and Green channels of the image ``kodim08'' from the Kodak PhotoCD Dataset, its synthetic noisy version, and the images recovered by the concatenated WNNM and the proposed MC-WNNM methods.}
\label{f1}
\vspace{-3mm}
\end{figure*}

The color image denoising problem is to recover the clean image $\mathbf{x}_{c}$ from its noisy version $\mathbf{y}_{c}=\mathbf{x}_{c}+\mathbf{n}_{c}$, where $c=\{r, g, b\}$ is the index of R, G, B channels and $\mathbf{n}_{c}$ is the noise in the $c$ channel.\ Patch based image denoising \cite{foe,ksvd,bm3d,lssc,epll,ncsr,mlp,csf,wnnm,pgpd,chen2015learning,dncnn} has achieved a great success in the last decade.\ Given a noisy color image $\mathbf{y}_{c}$, each local patch of size $p\times p \times 3$ is extracted and stretched to a patch vector, denoted by $\mathbf{y}=[\mathbf{y}_{r}^{\top}\ \mathbf{y}_{g}^{\top}\ \mathbf{y}_{b}^{\top}]^{\top}\in\mathbb{R}^{3p^{2}}$, where $\mathbf{y}_{r}, \mathbf{y}_{g}, \mathbf{y}_{b}\in\mathbb{R}^{p^{2}}$ are the corresponding patches in R, G, B channels.\ For each local patch $\mathbf{y}$, we search the $M$ most similar patches to it (including $\mathbf{y}$ itself) by Euclidean distance in a relatively large local window around it.\ By stacking the $M$ similar patches column by column, we form a noisy patch matrix $\mathbf{Y}=\mathbf{X}+\mathbf{N}\in\mathbb{R}^{3p^{2}\times M}$, where $\mathbf{X}$ and $\mathbf{N}$ are the corresponding clean and noise patch matrices.

The noise in standard RGB (sRGB) space could be approximately modeled as additive white Gaussian (AWGN), but noise in different channels has different variances \cite{Liu2008,Leungtip,crosschannel2016}.\ Therefore, it is problematic to directly apply some grayscale denoising methods to the concatenated vectors $\mathbf{y}$ or matrices $\mathbf{Y}$.\ To better illustrate this point, in Fig.\ \ref{f1}, we show a clean image ``kodim08'' (only the R and G channels are shown due to limit of space), its noisy version generated by adding AWGN to each channel, and the denoised image by applying WNNM \cite{wnnmijcv} to the concatenated  patch matrix $\mathbf{Y}$.\ The standard deviations of AWGN added to the R, G, B channels are $\sigma_{r}=40$, $\sigma_{g}=20$, $\sigma_{b}=30$, respectively.\ To make WNNM applicable to color image denoising, we set the noise standard deviation as the average deviation of the whole noisy image, i.e., $\sigma=\sqrt{(\sigma_{r}^{2}+\sigma_{g}^{2}+\sigma_{b}^{2})/3}\approx31.1$.\ From Fig.\ \ref{f1}, one can see that the concatenated WNNM remains some noise in the R channel while over-smoothing the G channel.\ This is because it processes R and G channels equally without considering their differences in noise corruption.

Clearly, a more effective color image denoising algorithm should consider the different noise strength in color channels.\ To this end, we introduce a weight matrix $\mathbf{W}$ to balance the noise in the RGB channels, and present the following multi-channel WNNM (MC-WNNM) model:
\vspace{-3mm}
\begin{equation}
\label{e4}
\vspace{-3mm}
\min_{\mathbf{X}}\|\mathbf{W}(\mathbf{Y}-\mathbf{X})\|_{F}^{2}
+ 
\|\mathbf{X}\|_{\bm{w},*}.
\end{equation}
We follow the method in \cite{wnnmijcv} to set the weight vector $\bm{w}$ on nuclear norm as $w_{i}^{k+1}=C/(|\sigma_{i}(\mathbf{X}_{k})|+\epsilon)$, where $\epsilon>0$ is a small number to avoid zero numerator and $\sigma_{i}(\mathbf{X}_{k})$ is the $i$th singular value of the estimated data matrix $\mathbf{X}$ at the $k$th iteration.\ Note that if $\sigma_{r}=\sigma_{g}=\sigma_{b}$, the proposed MC-WNNM model will be reduced to the concatenated WNNM model.\ With an appropriate setting of the weight matrix $\mathbf{W}$ and a good optimization algorithm, the proposed MC-WNNM model will lead to much better color image denoising results.\ As shown in Figs.\ \ref{f1}(f) and \ref{f1}(h), MC-WNNM removes clearly the noise in the R channel while preserving textures effectively in the G channel. 

\vspace{-1mm}
\subsection{The Setting of Weight Matrix $\mathbf{W}$}
\vspace{-1mm}

Let's denote the noisy patch matrix by $\mathbf{Y}=[\mathbf{Y}_{r}^{\top}\ \mathbf{Y}_{g}^{\top}\ \mathbf{Y}_{b}^{\top}]^{\top}$, where $\mathbf{Y}_{r}, \mathbf{Y}_{g}, \mathbf{Y}_{b}$ are sub-matrices of similar patches in R, G, B channels, respectively.\ The corresponding clean matrix is $\mathbf{X}=[\mathbf{X}_{r}^{\top}\ \mathbf{X}_{g}^{\top}\ \mathbf{X}_{b}^{\top}]^{\top}$, where $\mathbf{X}_{r}, \mathbf{X}_{g}, \mathbf{X}_{b}$ are similarly defined.\ The weight matrix $\mathbf{W}$ can be determined under the \emph{maximum a-posterior} (MAP) estimation framework:
\vspace{-3mm}
\begin{equation}
\label{e5}
\vspace{-3mm}
\begin{split}
\hat{\mathbf{X}} 
&=
\arg\max_{\mathbf{X}}\ln P(\mathbf{X}|\mathbf{Y},\bm{w})
\\
&
=
\arg\max_{\mathbf{X}}\{\ln P(\mathbf{Y}|\mathbf{X})+\ln P(\mathbf{X}|\bm{w})\}.
\end{split}
\end{equation}
The log-likelihood term $\ln P(\mathbf{Y}|\mathbf{X})$ is characterized by the
statistics of noise.\ According to \cite{Leungtip}, we assume that the noise is independent among RGB channels and independently and identically distributed (i.i.d.)\ in each channel with Gaussian distribution and standard deviations $\{\sigma_{r}, \sigma_{g}, \sigma_{b}\}$.\ There is:
\vspace{-3mm}
\begin{equation}
\label{e6}
\vspace{-3mm}
\hspace{-3mm}
P(\mathbf{Y}|\mathbf{X})
= 
\hspace{-4mm}
\prod_{c\in\{r, g, b\}}
\hspace{-4mm}
(2\pi\sigma_{c}^{2})^{-\frac{3p^{2}}{2}}
\exp(-\frac{1}{2\sigma_{c}^{2}}\|\mathbf{Y}_{c}-\mathbf{X}_{c}\|_{F}^{2}).
\end{equation}

For the latent data $\mathbf{X}$, the small weighted nuclear norm prior is imposed on it, i.e., $\|\mathbf{X}\|_{\bm{w},*}=\sum_{i}w_{i}\sigma_{i}(\mathbf{X})$ should be sparsely distributed.\ We let it be:
\vspace{-4mm}
\begin{equation}
\label{e7}
\vspace{-3mm}
P(\mathbf{X}|\bm{w})
\propto
\exp(-\frac{1}{2}\|\mathbf{X}\|_{\bm{w},*}).
\end{equation}
Putting (\ref{e7}) and (\ref{e6}) into (\ref{e5}), we have
\vspace{-3mm}
\begin{equation}
\label{e8}
\begin{split}
\vspace{-3mm}
\hat{\mathbf{X}}
&
=
\arg\min_{\mathbf{X}}
\sum_{c\in\{r, g, b\}}
\frac{1}{\sigma_{c}^{2}}\|(\mathbf{Y}_{c}-\mathbf{X}_{c})\|_{F}^{2}+\|\mathbf{X}\|_{\bm{w},*}
\\
&
=
\arg\min_{\mathbf{X}}\|\mathbf{W}(\mathbf{Y}-\mathbf{X})\|_{F}^{2}+\|\mathbf{X}\|_{\bm{w},*},
\end{split}
\end{equation}
with
\vspace{-2mm}
\begin{equation}
\label{e9}
\vspace{-3mm}
\mathbf{W}
=
\left( \begin{array}{ccc}
\sigma_{r}^{-1}\mathbf{I} & \mathbf{0} & \mathbf{0}
\\
\mathbf{0} & \sigma_{g}^{-1}\mathbf{I} & \mathbf{0}
\\
\mathbf{0} & \mathbf{0} & \sigma_{b}^{-1}\mathbf{I}
\end{array} \right),
\end{equation}
where $\mathbf{I}
\in\mathbb{R}^{p^{2}\times p^{2}}$ is the identity matrix. 

Clearly, the weight matrix $\mathbf{W}$ is diagonal and determined by the noise standard deviation in each channel.\ The stronger the noise in a channel, the less the contribution that channel should make to the estimation of $\mathbf{X}$.\ Our experimental results (refer to Section 4 please) on synthetic and real noisy images clearly demonstrate the advantages of MC-WNM over WNNM and other methods in color image denoising. 

\subsection{Model Optimization}

The proposed MC-WNNM model does not have an analytical solution.\ In the WNNM model \cite{wnnmijcv}, when the weights assigned on singular values are in a non-descending order, the weighted nuclear norm proximal operator can have a global optimum with closed-form solution.\ Unfortunately, such a property is not valid for the MC-WNNM model because a weight matrix $\mathbf{W}$ is assigned to the rows of data matrix $\mathbf{X}$.\ This makes the proposed model more difficult to solve than the original WNNM model.

We employ the variable splitting method \cite{courant1943,Eckstein1992} to solve the MC-WNNM model.\ By introducing an augmented variable $\mathbf{Z}$, the MC-WNNM model can be reformulated as a linear equality-constrained problem with two variables $\mathbf{X}$ and $\mathbf{Z}$:
\vspace{-3mm}
\begin{equation}
\label{e10}
\vspace{-3mm}
\min_{\mathbf{X},\mathbf{Z}}\|\mathbf{W}(\mathbf{Y}-\mathbf{X})\|_{F}^{2}
+
\|\mathbf{Z}\|_{\bm{w},*}
\quad
\text{s.t.}
\quad
\mathbf{X}=\mathbf{Z}.
\end{equation}
Since the objective function is separable w.r.t. the two variables, the problem (\ref{e10}) can be solved under the alternating direction method of multipliers (ADMM) \cite{admm} framework.\ The augmented Lagrangian function is:
\vspace{-2mm}
\begin{equation}
\label{e11}
\vspace{-3mm}
\begin{split}
\mathcal{L}(\mathbf{X},\mathbf{Z},\mathbf{A},\rho)
=
&\|\mathbf{W}(\mathbf{Y}-\mathbf{X})\|_{F}^{2}
+
\|\mathbf{Z}\|_{\bm{w},*}
\\
&
+
\langle
\mathbf{A},\mathbf{X}-\mathbf{Z}
\rangle
+
\frac{\rho}{2}
\|\mathbf{X}-\mathbf{Z}\|_{F}^{2},
\end{split}
\end{equation}
where $\mathbf{A}$ is the augmented Lagrangian multiplier and $\rho>0$ is the penalty parameter.\ We initialize the matrix variables $\mathbf{X}_{0}$, $\mathbf{Z}_{0}$, and $\mathbf{A}_{0}$ to be zero matrix and $\rho_{0}>0$ to be a suitable value.\ Denote by ($\mathbf{X}_{k}, \mathbf{Z}_{k}$) and $\mathbf{A}_{k}$ the optimization variables and Lagrange multiplier at iteration $k$ ($k=0,1,2,...$), respectively.\ By taking derivatives of the Lagrangian function $\mathcal{L}$ w.r.t. $\mathbf{X}$ and $\mathbf{Z}$ and setting the derivative function to be zero, we can alternatively update the variables as follows:
\vspace{2mm}
\\
(1) \textbf{Update $\mathbf{X}$ while fixing $\mathbf{Z}$ and $\mathbf{A}$}:
\vspace{-2mm}
\begin{equation}
\label{e12}
\mathbf{X}_{k+1}
=
\arg\min_{\mathbf{X}}
\|\mathbf{W}(\mathbf{Y}-\mathbf{X})\|_{F}^{2} 
+
\frac{\rho_{k}}{2}\|\mathbf{X} - \mathbf{Z}_{k} + \rho_{k}^{-1}\mathbf{A}_{k}||_{F}^{2}.
\end{equation}
This is a standard least squares regression problem with closed-form solution:
\vspace{-3mm}
\begin{equation}
\label{e13}
\vspace{-5mm}
\mathbf{X}_{k+1}
=
(\mathbf{W}^{\top}\mathbf{W}+\frac{\rho_{k}}{2}\mathbf{I})^{-1}
(\mathbf{W}^{\top}\mathbf{W}\mathbf{Y} + \frac{\rho_{k}}{2}\mathbf{Z}_{k} -\frac{1}{2}\mathbf{A}_{k}).
\end{equation}
\\
(2) \textbf{Update $\mathbf{Z}$ while fixing $\mathbf{X}$ and $\mathbf{A}$}:
\vspace{-3mm}
\begin{equation}
\label{e14}
\mathbf{Z}_{k+1}
=
\arg\min_{\mathbf{Z}}\frac{\rho_{k}}{2}
\|\mathbf{Z} - (\mathbf{X}_{k+1}+\rho_{k}^{-1}\mathbf{A}_{k})\|_{F}^{2}
+
\|\mathbf{Z}\|_{\bm{w},*}.
\end{equation}
According to the Theorem 1 in \cite{wnnmijcv}, given the SVD of $\mathbf{X}_{k+1}+\rho_{k}^{-1}\mathbf{A}_{k}$, i.e., $\mathbf{X}_{k+1}+\rho_{k}^{-1}\mathbf{A}_{k}=\mathbf{U}_{k}\mathbf{\Sigma}_{k}\mathbf{V}_{k}^{\top}$, where 
$\mathbf{\Sigma}_{k}=
\left( \begin{array}{c}
\text{diag}(\sigma_{1},\sigma_{2},...,\sigma_{M})
\\
\mathbf{0}
\end{array} \right)
\in\mathbb{R}^{3p^{2}\times M}$ (without loss of generality, we assume that $3p^{2}\ge M$),
the global optimum of the above problem is 
$\hat{\mathbf{Z}}_{k+1}=\mathbf{U}_{k}\hat{\mathbf{\Sigma}}_{k}\mathbf{V}_{k}^{\top}$, where 
$\hat{\mathbf{\Sigma}}_{k}=
\left( \begin{array}{c}
\text{diag}(\hat{\sigma}_{1},\hat{\sigma}_{2},...,\hat{\sigma}_{M})
\\
\mathbf{0}
\end{array} \right)
\in\mathbb{R}^{3p^{2}\times M}$
and $(\hat{\sigma}_{1},\hat{\sigma}_{2},...,\hat{\sigma}_{M})$ is the solution to the following convex optimization problem:
\vspace{-3mm}
\begin{equation}
\label{e15}
\vspace{-3mm}
\begin{split}
\min_{\hat{\sigma}_{1},\hat{\sigma}_{2},...,\hat{\sigma}_{M}}
&
\sum\nolimits_{i=1}^{M}
(\sigma_{i}-\hat{\sigma}_{i})^{2}
+
\frac{2w_{i}}{\rho_{k}}\hat{\sigma}_{i}
\\
&
\text{s.t.}
\quad
\hat{\sigma}_{1}\ge \hat{\sigma}_{2} \ge...\ge \hat{\sigma}_{M}\ge 0.
\end{split}
\end{equation}
According to the Remark 1 in \cite{wnnmijcv}, the problem above has closed-form solution ($i=1,2,...,M$):
\vspace{-3mm}
\begin{equation}
\label{e16}
\vspace{-2mm}
\hat{\sigma}_{i}
=
\left\{ \begin{array}{ll}
0 & \textrm{if $c_{2}<0$}\\
\frac{c_{1}+\sqrt{c_{2}}}{2} & \textrm{if $c_{2}\ge 0$}
\end{array} \right.,
\end{equation}
where $c_{1}=\sigma_{i}-\epsilon$, $c_{2} = (\sigma_{i}-\epsilon)^{2}-\frac{8C}{\rho_{k}}$, $\epsilon>0$ is a small number, and $C$ is set as $\sqrt{2M}$ by experience in \cite{wnnmijcv}.
\vspace{1mm}
 \\
(3) \textbf{Update $\mathbf{A}$ while fixing $\mathbf{X}$ and $\mathbf{Z}$}:
\vspace{-3mm}
\begin{equation}
\label{e17}
\vspace{-7mm}
\mathbf{A}_{k+1}
=
\mathbf{A}_{k} + \rho_{k}(\mathbf{X}_{k+1}-\mathbf{Z}_{k+1}).
\end{equation}
\\
(4) \textbf{Update $\rho_{k}$}: $\rho_{k+1}= \mu * \rho_{k}$, where $\mu>1$.
\vspace{1mm}

The above alternative updating steps are repeated until the convergence condition is satisfied or the number of iterations exceeds a preset threshold.\ The convergence condition of the ADMM algorithm is: $\|\mathbf{X}_{k+1}-\mathbf{Z}_{k+1}\|_{F}\le \text{Tol}$, $\|\mathbf{X}_{k+1}-\mathbf{X}_{k}\|_{F}\le \text{Tol}$, and $\|\mathbf{Z}_{k+1}-\mathbf{Z}_{k}\|_{F}\le \text{Tol}$ are simultaneously satisfied, where $\text{Tol}>0$ is a small tolerance number.\ We summarize the updating procedures in Algorithm 1.\ The convergence analysis of the proposed Algorithm 1 is given in Theorem \ref{th1}.\ Note that since the weighted nuclear norm is non-convex in general, we employ an unbounded sequence of $\{\rho_{k}\}$ here to make sure that Algorithm 1 converges. 

\begin{table}
\vspace{-0mm}
\begin{tabular}{l}
\hline
\textbf{Algorithm 1}: Solve MC-WNNM via ADMM
\\
\hline
\textbf{Input:} Matrices $\mathbf{Y}$ and $\mathbf{W}$, $\mu>1$, $\text{Tol}>0$, $K_{1}$;
\\
\textbf{Initialization:} $\mathbf{X}_{0}=\mathbf{Z}_{0}=\mathbf{A}_{0}=\mathbf{0}$, $\rho_{0}>0$,
\\
\quad \quad \quad \quad \quad \quad \ \text{T} = \text{False}, $k=0$; 
\\
\textbf{While} (\text{T} == \text{false}) \textbf{do}
\\
1. Update $\mathbf{X}_{k+1}$ as 
\\
$\mathbf{X}_{k+1}
\hspace{-1mm}
=
\hspace{-1mm}
(\mathbf{W}^{\top}\mathbf{W}+\frac{\rho_{k}}{2}\mathbf{I})^{-1}
(\mathbf{W}^{\top}\mathbf{W}\mathbf{Y} + \frac{\rho_{k}}{2}\mathbf{Z}_{k} -\frac{1}{2}\mathbf{A}_{k})
$
\\
2. Update $\mathbf{Z}_{k+1}$ by solving the problem 
\\
\quad 
\quad
$
\min_{\mathbf{Z}}\frac{\rho_{k}}{2}
\|\mathbf{Z} - (\mathbf{X}_{k+1}+\rho_{k}^{-1}\mathbf{A}_{k})\|_{F}^{2}
+
\|\mathbf{Z}\|_{\bm{w},*}
$
\\
3. Update $\mathbf{A}_{k+1}$ as
$
\mathbf{A}_{k+1}
=
\mathbf{A}_{k} + \rho_{k}(\mathbf{X}_{k+1}-\mathbf{Z}_{k+1})
$
\\
4. Update $\rho_{k+1}= \mu * \rho_{k}$;
\\
5. $k \leftarrow k + 1$;
\\
\quad \textbf{if} (Convergence condition is satisfied) or ($k\ge K_{1}$)
\\
6.\quad \text{T} $\leftarrow$ \text{True}
\\
\quad \textbf{end if}
\\
\textbf{end while}
\\
\textbf{Output:} Matrices $\mathbf{X}$ and $\mathbf{Z}$.
\\
\hline
\end{tabular}
\vspace{-3mm}
\label{a1}
\end{table}

\vspace{-1mm}
\begin{theorem}
\label{th1}
Assume that the weights in $\bm{w}$ are in a non-descending order, the sequences $\{\mathbf{X}_{k}\}$, $\{\mathbf{Z}_{k}\}$, and $\{\mathbf{A}_{k}\}$ generated in Algorithm 1 satisfy:
\vspace{-3mm}
\begin{align}
&(a) \lim_{k \to \infty} \|\mathbf{X}_{k+1}-\mathbf{Z}_{k+1}\|_{F}=0;
\\
&(b) \lim_{k \to \infty} \|\mathbf{X}_{k+1}-\mathbf{X}_{k}\|_{F}=0;
\\
&(c) \lim_{k \to \infty} \|\mathbf{Z}_{k+1}-\mathbf{Z}_{k}\|_{F}=0.
\end{align}
\end{theorem}
\vspace{-5mm}
\begin{proof}
We give a sketch proof here and detailed proof of this theorem can be found in the supplementary file. 

We first prove that the sequence $\{\mathbf{A}_{k}\}$ generated by Algorithm 1 is upper bounded.\ Since $\{\rho_{k}\}$ is unbounded, i.e., $\lim_{k\to\infty}{\rho_{k}}=+\infty$, we can prove that the sequence of Lagrangian function $\{\mathcal{L}(\mathbf{X}_{k+1},\mathbf{Z}_{k+1},\mathbf{A}_{k},\rho_{k})\}$ is also upper bounded.\ Hence, both $\{\mathbf{W}(\mathbf{Y}-\mathbf{X}_{k})\}$ and $\{\mathbf{Z}_{k}\}$ are upper bounded.\ Then $\{\mathbf{X}_{k}\}$ is also upper bounded.\ According to Eq.\ (\ref{e17}), we can prove that 
$
\lim_{k \to \infty} 
\|
\mathbf{X}_{k+1}
-
\mathbf{Z}_{k+1}
\|_{F}
=
\lim_{k \to \infty} 
\rho_{k}^{-1}
\|
\mathbf{A}_{k+1}
-
\mathbf{A}_{k}
\|_{F}
=
0
$,
and (a) is proved.\ Then we can prove that 
$
\lim_{k \to \infty} 
\|
\mathbf{X}_{k+1}
-
\mathbf{X}_{k}
\|_{F}
\le
\lim_{k \to \infty} 
(\|
(\mathbf{W}^{\top}\mathbf{W}
+
\frac{\rho_{k}}{2}
\mathbf{I})^{-1}
(\mathbf{W}^{\top}\mathbf{W}\mathbf{Y}
-
\mathbf{W}^{\top}\mathbf{W}\mathbf{Z}_{k}
-
\frac{1}{2}
\mathbf{A}_{k})
\|_{F}
+
\rho_{k}^{-1}\|
\mathbf{A}_{k}-\mathbf{A}_{k-1}
\|_{F})
=
0
$
and hence (b) is proved.\ Finally, (c) can be proved by checking that 
$
\lim_{k \to \infty} \|\mathbf{Z}_{k+1}-\mathbf{Z}_{k}\|
\le
\lim_{k \to \infty} 
(\|
\mathbf{\Sigma}_{k-1}-\mathcal{S}_{\bm{w}/\rho_{k-1}}(\mathbf{\Sigma}_{k-1})
\|_{F}
+
\|
\mathbf{X}_{k+1}-\mathbf{X}_{k}
\|_{F}
+
\|
\rho_{k-1}^{-1}\mathbf{A}_{k-1}
+
\rho_{k}^{-1}\mathbf{A}_{k+1}
-
\rho_{k}^{-1}\mathbf{A}_{k}
\|_{F})
=
0,
$
where $\mathbf{U}_{k-1}\mathbf{\Sigma}_{k-1}\mathbf{V}_{k-1}^{\top}$ is the SVD of $\mathbf{X}_{k}+\rho_{k-1}^{-1}\mathbf{A}_{k-1}$
.
\vspace{-2mm}
\end{proof}

\vspace{-2mm}
\subsection{The Denoising Algorithm}
\vspace{-2mm}

Given a noisy color image $\mathbf{y}_{c}$, suppose that we have extracted $N$ local patches $\{\mathbf{y}_{j}\}_{j=1}^{N}$ and their similar patches.\ $N$ noisy patch matrices $\{\mathbf{Y}_{j}\}_{j=1}^{N}$ can be formed to estimate the clean matrices $\{\mathbf{X}_{j}\}_{j=1}^{N}$.\ The patches in matrices $\{\mathbf{X}_{j}\}_{j=1}^{N}$ are aggregated to form the denoised  image $\hat{\mathbf{x}}_{c}$.\ To obtain better denoising results, we perform the above denoising procedures for several rounds.\ The proposed MC-WNNM based color image denoising algorithm is summarized in Algorithm 2.
\begin{table}
\vspace{-0mm}
\begin{tabular}{l}
\hline
\textbf{Algorithm 2}: Color Image Denoising by MC-WNNM
\\
\hline
\textbf{Input:} Noisy image $\mathbf{y}_{c}$, noise levels $\{\sigma_{r}, \sigma_{g}, \sigma_{b}\}$, $K_{2}$;
\\
\textbf{Initialization:} $\hat{\mathbf{x}}_{c}^{(0)}=\mathbf{y}_{c}$, $\mathbf{y}_{c}^{(0)}=\mathbf{y}_{c}$;
\\
\textbf{for} $k = 1:K_{2}$ \textbf{do}
\\
1. Set $\mathbf{y}_{c}^{(k)}=\hat{\mathbf{x}}_{c}^{(k-1)}$;
\\
2. Extract local patches $\{\mathbf{y}_{j}\}_{j=1}^{N}$ from $\mathbf{y}_{c}^{(k)}$;
\\
\quad\textbf{for} each patch $\mathbf{y}_{j}$ \textbf{do}
\\
3.\quad Search non-local similar patches $\mathbf{Y}_{j}$;
\\
4.\quad Apply the MC-WNNM model (\ref{e10}) to $\mathbf{Y}_{j}$ and
\\
\quad \quad 
obtain the estimated $\mathbf{X}_{j}$;
\\
\quad\textbf{end for}
\\
5. Aggregate $\{\mathbf{X}_{j}\}_{j=1}^{N}$ to form the image $\hat{\mathbf{x}}_{c}^{(k)}$;
\\
\textbf{end for}
\\
\textbf{Output:} Denoised image $\hat{\mathbf{x}}_{c}^{(K_{2})}$.
\\
\hline
\end{tabular}
\vspace{-2mm}
\label{a2}
\end{table}

\vspace{-1mm}
\subsection{Complexity Analysis}
\vspace{-1mm}

In Algorithm 1 for solving the MC-WNNM model via ADMM, the cost for updating $\bm{X}$ is $\mathcal{O}(\max(p^{4}M, M^{3}))$, while the cost for updating $\bm{Z}$ is $\mathcal{O}(p^{4}M+M^{3})$.\ The costs for updating $\bm{A}$ and $\rho$ can be ignored.\ So the overall complexity is $\mathcal{O}((p^{4}M+M^{3})K_{1})$, where $K_{1}$ is the number of iterations.\ In Algorithm 2 for image denoising, we consider the number of patches $N$ extracted from the input noisy image and the number of itertations $K_{2}$ and ignore the cost for searching similar patches.\ The overall cost is $\mathcal{O}((p^{4}M+M^{3})K_{1}K_{2}N)$.

\vspace{-1mm}
\section{Experiments}
\vspace{-1mm}

We evaluate the proposed MC-WNNM method on synthetic and real noisy color images.\ We compare the proposed method with state-of-the-art denoising methods, including CBM3D \cite{cbm3d}, MLP \cite{mlp}, WNNM \cite{wnnm}, TNRD \cite{chen2015learning}, DnCNN \cite{dncnn} ``Noise Clinic'' (NC) \cite{noiseclinic,ncwebsite}, CC \cite{crosschannel2016}, and the commercial software Neat Image (NI) \cite{neatimage}. The Matlab source code of our MC-WNNM algorithm can be downloaded at \url{http://www4.comp.polyu.edu.hk/~cslzhang/code/MCWNNM.zip}.

\vspace{-1mm}
\subsection{Experimental Settings}
\vspace{-1mm}

\textbf{Noise level estimation.}\
For most of the competing denoising algorithms, the standard deviation of noise should be given as a parameter.\ In synthetic experiments, the noise levels ($\sigma_{r}, \sigma_{g}, \sigma_{b}$) in R, G ,B channels are assumed to be known.\ In the case of real noisy images, the noise levels can be estimated via some noise estimation methods \cite{noiselevel,Chen2015ICCV}.\ In this paper, we employ the method \cite{Chen2015ICCV} to estimate the noise level for each color channel. 

\textbf{Noise level of comparison methods.}\ 
For the CBM3D method \cite{cbm3d}, a single parameter of noise level should be input. We set the noise level as
\vspace{-3mm}
\begin{equation}
\label{e21}
\vspace{-3mm}
\sigma = \sqrt{(\sigma_{r}^{2}+\sigma_{g}^{2}+\sigma_{b}^{2})/3}.
\end{equation}
The methods of MLP \cite{mlp} and TNRD \cite{chen2015learning} are originally designed for grayscale images.\ We retrain their models (using the released codes by the authors) at different noise levels from $\sigma=5$ to $\sigma=75$ with a gap of $5$.\ The denoising on color images is performed by processing each channel with the model trained at the same (or nearest) noise level. 

\textbf{Comparison with WNNM.}\
In order to make a full and fair comparison with the original WNNM method \cite{wnnmijcv}, we implement WNNM for color image denoising in three ways.\ 
1) We apply WNNM to each color channel separately with the corresponding noise levels $\sigma_{r}, \sigma_{g}, \sigma_{b}$.\ We call this method ``WNNM-1''.\
2) We perform WNNM on the concatenated matrix $\mathbf{Y}$ formed by the patches in RGB channels, while the input noise level $\sigma$ is computed by Eq.\ (\ref{e21}).\ We call this method ``WNNM-2''.\
3) We set the weight matrix $\mathbf{W}$ as $\mathbf{W}=\sigma^{-1}\mathbf{I}$ in the proposed MC-WNNM model, and use our developed algorithm for denoising.\ We call this method ``WNNM-3''.

For a fair comparison, we tune the parameters of WNNM-1, WNNM-2, WNNM-3 and MC-WNNM to achieve their best denoising performance.\ The detailed parameters are as follows: we set the patch size as $p = 6$, the number of non-local similar patches as $M = 70$, the window size for searching similar patches as $40\times40$.\ For WNNM-3 and MC-WNNM, the updating parameter is set as $\mu=1.001$.\ The number of iterations in Algorithm 1 is set as $K_{1} = 10$.\ The number of iterations $K_{2}$ in Algorithm 2 and the initial penalty parameter $\rho_{0}$ will be given in the following sub-sections.

\begin{figure}
\vspace{-3mm}
\centering
\subfigure{
\begin{minipage}{0.08\textwidth}
\includegraphics[width=1\textwidth]{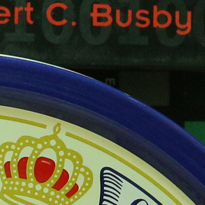}
\end{minipage}
\begin{minipage}{0.08\textwidth}
\includegraphics[width=1\textwidth]{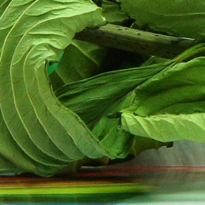}
\end{minipage}
\begin{minipage}{0.08\textwidth}
\includegraphics[width=1\textwidth]{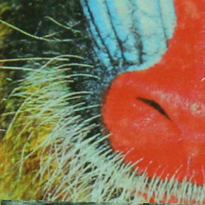}
\end{minipage}
\begin{minipage}{0.08\textwidth}
\includegraphics[width=1\textwidth]{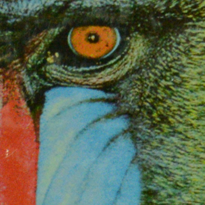}
\end{minipage}
\begin{minipage}{0.08\textwidth}
\includegraphics[width=1\textwidth]{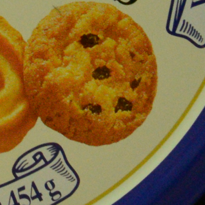}
\end{minipage}
}\vspace{-3mm}
\subfigure{
\begin{minipage}{0.08\textwidth}
\includegraphics[width=1\textwidth]{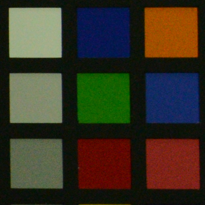}
\end{minipage}
\begin{minipage}{0.08\textwidth}
\includegraphics[width=1\textwidth]{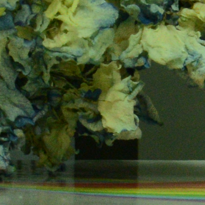}
\end{minipage}
\begin{minipage}{0.08\textwidth}
\includegraphics[width=1\textwidth]{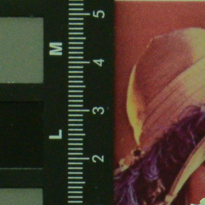}
\end{minipage}
\begin{minipage}{0.08\textwidth}
\includegraphics[width=1\textwidth]{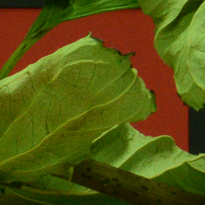}
\end{minipage}
\begin{minipage}{0.08\textwidth}
\includegraphics[width=1\textwidth]{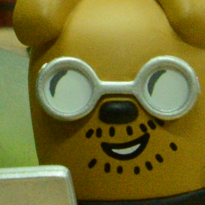}
\end{minipage}
}\vspace{-3mm}
\subfigure{
\begin{minipage}{0.08\textwidth}
\includegraphics[width=1\textwidth]{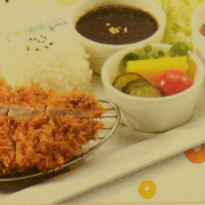}
\end{minipage}
\begin{minipage}{0.08\textwidth}
\includegraphics[width=1\textwidth]{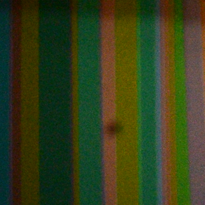}
\end{minipage}
\begin{minipage}{0.08\textwidth}
\includegraphics[width=1\textwidth]{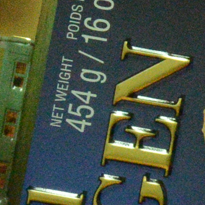}
\end{minipage}
\begin{minipage}{0.08\textwidth}
\includegraphics[width=1\textwidth]{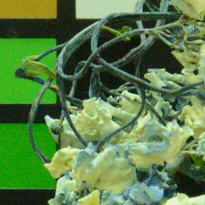}
\end{minipage}
\begin{minipage}{0.08\textwidth}
\includegraphics[width=1\textwidth]{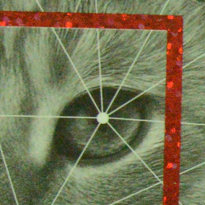}
\end{minipage}
}\vspace{-1mm}
\caption{The 15 cropped real noisy images used in \cite{crosschannel2016}.}
\label{f2}
\vspace{-5mm}
\end{figure}

\vspace{-1mm}
\subsection{Experiments on Synthetic Noisy Color Images}
\vspace{-1mm}

We first compare MC-WNNM with the competing denoising methods \cite{cbm3d,mlp,chen2015learning,dncnn,noiseclinic,neatimage} on the 24 color images from the Kodak PhotoCD Dataset (\url{http://r0k.us/graphics/kodak/}).\ The noisy images are generated by adding AWGN to each of the R, G, B channels, respectively.\ In the main paper, we report the results by setting $\sigma_{r}=40, \sigma_{g}=20, \sigma_{b}=30$.\ More results with other noise settings can be found in the supplementary file.\ For WNNM-3 and MC-WNNM, the initial penalty parameter is set as $\rho_{0}=10$ and $\rho_{0}=3$, respectively.\ The number of iterations in Algorithm 2 is set as $K_{2}=8$. 

The PSNR results by competing methods are listed in Table \ref{t1}, while the best PSNR result for each image is highlighted in bold.\ One can see that on all the 24 images, our method achieves the highest PSNR values among the competing methods.\ On average, MC-WNNM achieves 0.47dB, 0.48dB and 1.09dB improvements over WNNM-1, WNNM-2 and WNNM-3, respectively.\ For space limitation, we leave the visual comparisons of the synthetic noisy image denoising results in the supplementary file.

\begin{table*}
\vspace{-6mm}
\caption{PSNR(dB) results of different denoising methods on 24 natural color images.}
\vspace{0.5mm}
\label{t1}
\label{taba}
\begin{center}
\renewcommand\arraystretch{1.0}
\scriptsize
\begin{tabular}{|c||c|c|c|c|c|c|c|c|c|c|}
\hline
&\multicolumn{10}{c|}{ $\sigma_{r} = 40, \sigma_{g} = 20, \sigma_{b} = 30$}
\\
\hline
\hline
Image\#
&
\textbf{CBM3D}
&
\textbf{MLP}
&
\textbf{TNRD}
&
\textbf{DnCNN}
&
\textbf{NI}
&
\textbf{NC}
&
\textbf{WNNM-1}
&
\textbf{WNNM-2}
&
\textbf{WNNM-3}
&
\textbf{MC-WNNM}
\\
\hline
1& 25.24 & 25.70 & 25.74 & 20.47 & 23.85 & 24.90 & 26.01 & 25.95 & 25.58 & \textbf{26.66}
\\
\hline
2& 28.27 & 30.12 & 30.21 & 20.47 & 25.90 & 25.87 & 30.08 & 30.11 & 29.80 & \textbf{30.20} 
\\
\hline
3& 28.81 & 31.19 & 31.49 & 20.53 & 26.00 & 28.58 & 31.58 & 31.61 & 31.20 & \textbf{32.25}  
\\
\hline 
4& 27.95 & 29.88 & 29.86 & 20.47 & 25.82 & 25.67 & 30.13 & 30.16 & 29.84 & \textbf{30.49} 
\\
\hline
5& 25.03 & 26.00 & 26.18 & 20.52 & 24.38 & 25.15 & 26.44 & 26.39 & 25.32 & \textbf{26.82}
\\
\hline
6& 26.24 & 26.84 & 26.90 & 20.66 & 24.65 & 24.74 & 27.39 & 27.30 & 26.88 & \textbf{27.98} 
\\
\hline
7& 27.88 & 30.28 & 30.40 & 20.52 & 25.63 & 27.69 & 30.47 & 30.54 & 29.70 & \textbf{30.98} 
\\
\hline
8& 25.05 & 25.59 & 25.83 & 20.57 & 24.02 & 25.30 & 26.71 & 26.75 & 25.26 & \textbf{26.90}
\\
\hline
9& 28.44 & 30.75 & 30.81 & 20.50 & 25.94 & 27.44 & 30.86 & 30.92 & 30.29 & \textbf{31.49}
\\
\hline
10& 28.27 & 30.38 & 30.57 & 20.52 & 25.87 & 28.42 & 30.65 & 30.68 & 29.95 & \textbf{31.26}
\\
\hline
11& 26.95 & 28.00 & 28.14 & 20.52 & 25.32 & 24.67 & 28.19 & 28.16 & 27.61 & \textbf{28.63}
\\
\hline
12& 28.76 & 30.87 & 31.05 & 20.60 & 26.01 & 28.37 & 30.97 & 31.06 & 30.58 & \textbf{31.48}
\\
\hline
13& 23.76 & 23.95 & 23.99 & 20.52 & 23.53 & 22.76 & 24.27 & 24.15 & 23.52 & \textbf{24.89}
\\
\hline
14& 26.02 & 26.97 & 27.11 & 20.51 & 24.94  & 25.68 & 27.20 & 27.15 & 26.55 & \textbf{27.57}
\\
\hline
15& 28.38 & 30.15 & 30.44 & 20.71 & 26.06 & 28.21 & 30.52 & 30.60 & 30.13 & \textbf{30.81}
\\
\hline
16& 27.75 & 28.82 & 28.87 & 20.52 & 25.69 & 26.66 & 29.27 & 29.21 & 29.02 & \textbf{29.96}
\\
\hline
17& 27.90 & 29.57 & 29.80 & 20.56 & 25.85 & 28.32 & 29.78 & 29.79 & 29.16 & \textbf{30.40}
\\
\hline
18& 25.77 & 26.40 & 26.41 & 20.53 & 24.74 & 25.70 & 26.63 & 26.56 & 26.01 & \textbf{27.22}
\\
\hline
19& 27.30 & 28.67 & 28.81 & 20.53 & 25.40 & 26.52 & 29.19 & 29.22 & 28.67 & \textbf{29.57}
\\
\hline
20& 28.96 & 30.40 & 30.76 & 21.44 & 24.95 & 25.90 & 30.79 & 30.83 & 29.97 & \textbf{31.07}
\\
\hline
21& 26.54 & 27.53 & 27.60 & 20.51 & 25.06 & 26.48 & 27.80 & 27.75 & 27.12 & \textbf{28.34}
\\
\hline
22& 27.05 & 28.17 & 28.27 & 20.51 & 25.36 & 26.60 & 28.21 & 28.16 & 27.81 & \textbf{28.64}
\\
\hline
23& 29.14 & 32.31 & 32.51 & 20.54 & 26.13 & 23.24 & 31.89 & 31.97 & 31.21 & \textbf{32.34}
\\
\hline
24& 25.75 & 26.41 & 26.53 & 20.59 & 24.55 & 25.73 & 27.10 & 27.03 & 26.18 & \textbf{27.59}
\\
\hline
\textbf{Average} 
& 27.13 & 28.54 & 28.68 & 20.58 & 25.24 & 26.19 & 28.84 & 28.83 & 28.22 & \textbf{29.31}
\\
\hline
\end{tabular}
\end{center}
\vspace{-1mm}
\end{table*}

\vspace{-1mm}
\subsection{Experiments on Real Noisy Color Images}
\vspace{-1mm}

We evaluate the proposed method on two real noisy color image datasets, where the images were captured under indoor or outdoor lighting conditions by different types of cameras and camera settings.\ For WNNM-3 and MC-WNNM, the initial penalty parameter is set as $\rho_{0}=8$ and $\rho_{0}=6$, respectively.\ The number of iterations in Algorithm 2 is set as $K_{2}=2$.

The first dataset is provided in \cite{ncwebsite}, which includes 20 real noisy images collected under uncontrolled outdoor environment.\ Since there is no ``ground truth'' of the noisy images, the objective measures such as PSNR cannot be computed on this dataset.

The second dataset is provided in \cite{crosschannel2016}, which includes noisy images of 11 static scenes.\ The noisy images were collected under controlled indoor environment.\ Each scene was shot 500 times under the same camera and camera setting.\ The mean image of the 500 shots is roughly taken as the ``ground truth", with which the PSNR can be computed.\ Since the image size is very large (about $7000\times5000$) and the 11 scenes share repetitive contents, the authors of \cite{crosschannel2016} cropped 15 smaller images of size $512\times512$ for experiments.\ Fig.\ \ref{f2} shows the contents of these images.\ Quantitative comparisons on the 15 cropped images will be reported. 

\vspace{-2mm}
\subsubsection{Results on Dataset \cite{ncwebsite}}
\vspace{-2mm}

\quad Since there is no ``ground truth'' for the real noisy images in dataset \cite{ncwebsite}, we only compare the visual quality of the denoised images by the compared methods.\ (Note that the method CC \cite{crosschannel2016} is not compared here since its code is not publically available.)

\begin{figure*}[t]
\vspace{-1mm}
\centering
\subfigure{
\begin{minipage}[t]{0.19\textwidth}
\centering
\raisebox{-0.15cm}{\includegraphics[width=1\textwidth]{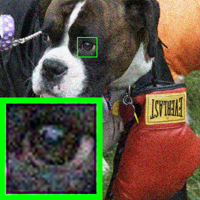}}
{\footnotesize (a) Noisy \cite{ncwebsite}   }
\end{minipage}
\begin{minipage}[t]{0.19\textwidth}
\centering
\raisebox{-0.15cm}{\includegraphics[width=1\textwidth]{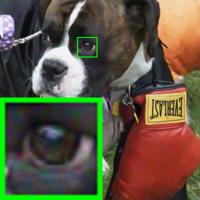}}
{\footnotesize (b) CBM3D \cite{cbm3d}  }
\end{minipage}
\begin{minipage}[t]{0.19\textwidth}
\centering
\raisebox{-0.15cm}{\includegraphics[width=1\textwidth]{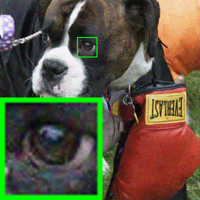}}
{\footnotesize (c) MLP \cite{mlp}  }
\end{minipage}
\begin{minipage}[t]{0.19\textwidth}
\centering
\raisebox{-0.15cm}{\includegraphics[width=1\textwidth]{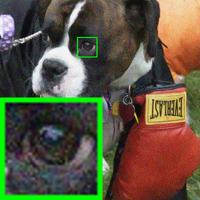}}
{\footnotesize (d) TNRD \cite{chen2015learning}}
\end{minipage}
\begin{minipage}[t]{0.19\textwidth}
\centering
\raisebox{-0.15cm}{\includegraphics[width=1\textwidth]{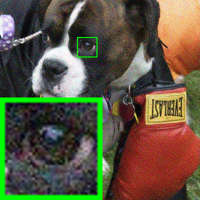}}
{\footnotesize (e) NI \cite{neatimage}  }
\end{minipage}
}\vspace{-3mm}
\subfigure{
\begin{minipage}[t]{0.19\textwidth}
\centering
\raisebox{-0.15cm}{\includegraphics[width=1\textwidth]{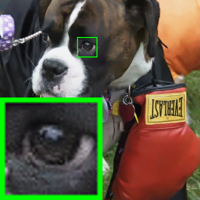}}
{\footnotesize (f) NC \cite{noiseclinic,ncwebsite}   }
\end{minipage}
\begin{minipage}[t]{0.19\textwidth}
\centering
\raisebox{-0.15cm}{\includegraphics[width=1\textwidth]{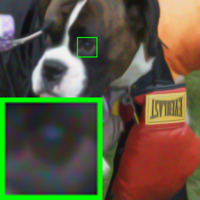}}
{\footnotesize (g) WNNM-1 \cite{wnnm}   }
\end{minipage}
\begin{minipage}[t]{0.19\textwidth}
\centering
\raisebox{-0.15cm}{\includegraphics[width=1\textwidth]{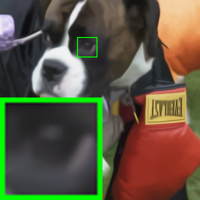}}
{\footnotesize (h) WNNM-2 \cite{wnnm}   }
\end{minipage}
\begin{minipage}[t]{0.19\textwidth}
\centering
\raisebox{-0.15cm}{\includegraphics[width=1\textwidth]{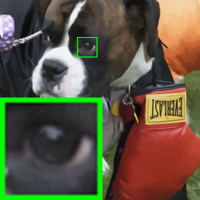}}
{\footnotesize (i) WNNM-3 \cite{wnnm}   }
\end{minipage}
\begin{minipage}[t]{0.19\textwidth}
\centering
\raisebox{-0.15cm}{\includegraphics[width=1\textwidth]{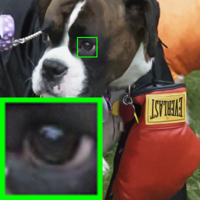}}
{\footnotesize (j) MC-WNNM  }
\end{minipage}
}
\vspace{-1mm}
\caption{Denoised images of the real noisy image ``Dog'' \cite{ncwebsite} by different methods.\ The estimated noise levels of R, G, and B channels are 16.8, 17.0, and 16.6, respectively.\ The images are better to be zoomed in on screen.}
\label{f3}
\vspace{-3mm}
\end{figure*}

Fig.\ \ref{f3} shows the denoised images of ``Dog'' by the competing methods.\ It can be seen that CBM3D, MLP, TRND and WNNM-1 tend to generate some noise caused color artifacts.\ Besides, WNNM-2 and WNNM-3 tend to over-smooth much the image.\ These results demonstrate that for color image denoising, neither processing each channel separately nor processing the three channels jointly but ignoring their noise difference is an effective solution.\ Though NC and NI methods are specifically developed for real color image denoising, their performance is not very satisfactory.\ In comparison, the proposed MC-WNNM recovers much better the structures and textures (such as the eye area) than the other competing methods.\ More visual comparisons on this dataset can be found in the supplementary file.

\begin{table*}
\caption{PSNR(dB) results and averaged computational time (s) of different methods on 15 cropped real noisy images used in \cite{crosschannel2016}.}
\vspace{0.5mm}
\label{t2}
\label{tabb}
\begin{center}
\renewcommand\arraystretch{1}
\scriptsize
\begin{tabular}{|c||c|c|c|c|c|c|c|c|c|c|c|}
\hline
Camera Settings  
&
\textbf{CBM3D}
&
\textbf{MLP}
&
\textbf{TNRD}
&
\textbf{DnCNN}
&
\textbf{NI}
&
\textbf{NC}
&
\textbf{CC}
&
\textbf{WNNM-1}
&
\textbf{WNNM-2}
&
\textbf{WNNM-3}
&
\textbf{MC-WNNM} 
\\
\hline
\multirow{3}{*}{\small{Canon 5D}} 
& 39.76 & 39.00 & 39.51 & 37.26 & 35.68 & 36.20 & 38.37 & 37.51 & 39.74 & 39.98 & \textbf{41.13}
\\ 
\cline{2-12} 
\multirow{3}{*}{ISO = 3200}   
& 36.40 & 36.34 & 36.47 & 34.13 & 34.03 & 34.35 & 35.37 & 33.86 & 35.12 & 36.65 & \textbf{37.28}
\\ 
\cline{2-12}    
& 36.37 & 36.33 & 36.45 & 34.09 & 32.63 & 33.10 & 34.91 & 31.43 & 33.14 & 34.63 & \textbf{36.52}  
\\
\hline
\multirow{3}{*}{Nikon D600} 
& 34.18 & 34.70 & 34.79 & 33.62 & 31.78 & 32.28 & 34.98 & 33.46 & 35.08 & 35.08 & \textbf{35.53}
\\ 
\cline{2-12} 
\multirow{3}{*}{ISO = 3200}   
& 35.07 & 36.20 & 36.37 & 34.48 & 35.16 & 35.34 & 35.95 & 36.09 & 36.42 & 36.84 & \textbf{37.02}
\\ 
\cline{2-12}    
& 37.13 & 39.33 & 39.49 & 35.41 & 39.98 & 40.51 & \textbf{41.15} & 39.86 & 40.78 & 39.24 & 39.56
\\
\hline
\multirow{3}{*}{Nikon D800} 
& 36.81  & 37.95 & 38.11 & 35.79 & 34.84 & 35.09 & 37.99 & 36.35 & 38.28 & 38.61 & \textbf{39.26}
\\ 
\cline{2-12} 
\multirow{3}{*}{ISO = 1600}   
& 37.76 & 40.23 & 40.52 & 36.08 & 38.42 & 38.65 & 40.36 & 39.99 & 41.24 & 40.81 & \textbf{41.43}
\\ 
\cline{2-12}    
& 37.51 & 37.94 & 38.17 & 35.48 & 35.79 & 35.85 & 38.30 & 37.15 & 38.04 & 38.96 & \textbf{39.55}
\\
\hline
\multirow{3}{*}{Nikon D800} 
& 35.05 & 37.55 & 37.69 & 34.08 & 38.36 & 38.56 & 39.01 & 38.60 & \textbf{39.93} & 37.97 & 38.91
\\ 
\cline{2-12} 
\multirow{3}{*}{ISO = 3200}   
& 34.07 & 35.91 & 35.90 & 33.70 & 35.53 & 35.76 & 36.75 & 36.04 & 37.32 & 37.30 & \textbf{37.41}
\\ 
\cline{2-12}    
& 34.42 & 38.15 & 38.21 & 33.31 & 40.05 & 40.59 & 39.06 & 39.73 & \textbf{41.52} & 38.68 & 39.39
\\ 
\hline
\multirow{3}{*}{Nikon D800} 
& 31.13 & 32.69 & 32.81 & 29.83 & 34.08 & 34.25 & 34.61 & 33.29 & \textbf{35.20} & 34.57 & 34.80
\\ 
\cline{2-12} 
\multirow{3}{*}{ISO = 6400}   
& 31.22 & 32.33 & 32.33 & 30.55 & 32.13 & 32.38  & 33.21 & 31.16 & 33.61 & 33.43 & \textbf{33.95}
\\ 
\cline{2-12}    
& 30.97 & 32.29 & 32.29 & 30.09 & 31.52 & 31.76 & 33.22 & 31.98 & 33.62 & \textbf{34.02} & 33.94
\\
\hline
\textbf{Average} 
& 35.19 & 36.46 & 36.61 & 33.86 & 35.33 & 35.65 & 36.88 & 35.77 & 37.27 & 37.12 & \textbf{ 37.71}
\\
\hline
\textbf{Time} 
& 7.8 & 20.4 & 6.7 & 180.3 & \textbf{0.9} & 18.2 & NA & 689.1 & 465.3 & 198.6 & 202.9
\\
\hline
\end{tabular}
\end{center}
\vspace{-0mm}
\end{table*}

\begin{figure*}
\vspace{-0mm}
\centering
\subfigure{
\begin{minipage}[t]{0.19\textwidth}
\centering
\raisebox{-0.15cm}{\includegraphics[width=1\textwidth]{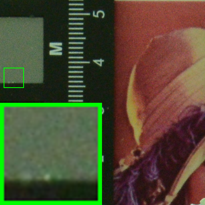}}
{\footnotesize (a) Noisy  \cite{crosschannel2016}: 35.71dB }
\end{minipage}
\begin{minipage}[t]{0.19\textwidth}
\centering
\raisebox{-0.15cm}{\includegraphics[width=1\textwidth]{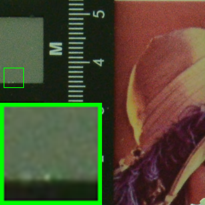}}
{\footnotesize (b) CBM3D \cite{bm3d,cbm3d}: 37.76dB}
\end{minipage}
\begin{minipage}[t]{0.19\textwidth}
\centering
\raisebox{-0.15cm}{\includegraphics[width=1\textwidth]{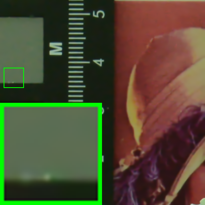}}
{\footnotesize (c) TNRD \cite{chen2015learning}: 40.52dB  } 
\end{minipage}
\begin{minipage}[t]{0.19\textwidth}
\centering
\raisebox{-0.15cm}{\includegraphics[width=1\textwidth]{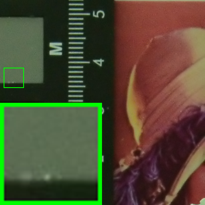}}
{\footnotesize (d) NI \cite{neatimage}: 38.42dB  }
\end{minipage}
\centering
\begin{minipage}[t]{0.19\textwidth}
\raisebox{-0.15cm}{\includegraphics[width=1\textwidth]{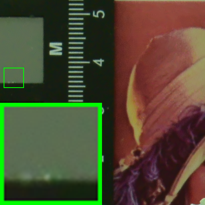}}
{\footnotesize (e) NC \cite{ncwebsite,noiseclinic}: 38.65dB  }
\end{minipage}
}\vspace{-3mm}
\subfigure{
\begin{minipage}[t]{0.19\textwidth}
\centering
\raisebox{-0.15cm}{\includegraphics[width=1\textwidth]{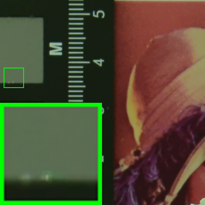}}
{\footnotesize (f) CC \cite{crosschannel2016}: 40.36dB }
\end{minipage}
\begin{minipage}[t]{0.19\textwidth}
\centering
\raisebox{-0.15cm}{\includegraphics[width=1\textwidth]{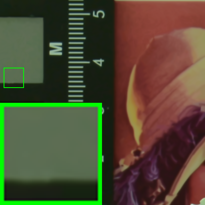}}
{\footnotesize (g) WNNM-2: 41.24dB}
\end{minipage}
\begin{minipage}[t]{0.19\textwidth}
\centering
\raisebox{-0.15cm}{\includegraphics[width=1\textwidth]{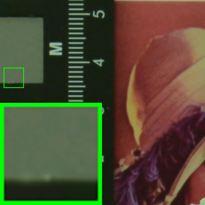}}
{\footnotesize (h) WNNM-3: 40.81dB}
\end{minipage}
\begin{minipage}[t]{0.19\textwidth}
\centering
\raisebox{-0.15cm}{\includegraphics[width=1\textwidth]{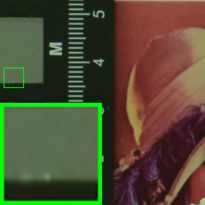}}
{\footnotesize (i) MC-WNNM: \textbf{41.43}dB}
\end{minipage}
\begin{minipage}[t]{0.19\textwidth}
\centering
\raisebox{-0.15cm}{\includegraphics[width=1\textwidth]{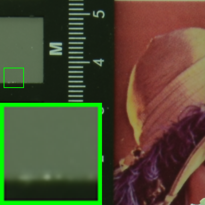}}
{\footnotesize (j) Mean Image \cite{crosschannel2016}}
\end{minipage}
}\vspace{-1mm}
\caption{Denoised images of a region cropped from the real noisy image ``Nikon D800 ISO=1600 2" \cite{crosschannel2016} by different methods.\ The estimated noise levels of R, G, and B channels are 1.3, 1.1, and 1.4, respectively.\ The images are better to be zoomed in on screen.}
\vspace{-2mm}
\label{f4}
\end{figure*}
\vspace{-2mm}
\subsubsection{Results on Dataset \cite{crosschannel2016}}
\vspace{-2mm}

\quad As described at the beginning of Section 4.3, there is a mean image for each noisy image in dataset \cite{crosschannel2016}, and those mean images can be roughly taken as ``ground truth'' for quantitative evaluation of denoising algorithms.

The results on PSNR and averaged computational time by competing methods (including CC \cite{crosschannel2016} whose results are copied from \cite{crosschannel2016}) are listed in Table \ref{t2}.\ For methods MLP \cite{mlp} and TNRD \cite{chen2015learning}, both of them achieve the best results when setting the noise level of the trained models at $\sigma=10$.\ The highest PSNR results are highlighted in bold.\ On average, MC-WNNM achieves 1.94dB, 0.44dB, 0.59dB improvements over the three WNNM methods, and significantly outperforms other competing method, including CC \cite{crosschannel2016}.\ On 10 out of the 15 images, the proposed MC-WNNM achieves the highest PSNR values, while WNNM-2 achieves the highest PSNR results on 3 of 15 images.\ It should be noted that in the CC method \cite{crosschannel2016}, a specific model is trained for each camera and camera setting, while the other methods uses the same model for all cases. 

Fig.\ \ref{f4} shows the denoised images of a scene captured by Nikon D800 ISO=1600.\ (The results of DnCNN and WNNM-1 are not shown here due to the limit of space.)\ We can see that CBM3D, NI, NC and CC will either remain noise or generate color artifacts, while TNRD, WNNM-2 and WNNM-3 over-smooth the image.\ In addition, due to treating each channel equally, both the denoised images (Fig.\ \ref{f4}(g) and Fig.\ \ref{f4}(h)) by WNNM-2 and WNNM-3 have chromatic aberration compared to the mean image (Fig.\ \ref{f4}(j)).\ MC-WNNM results in much better visual quality than other methods.\ More visual comparisons can be found in the supplementary file.

\textbf{Comparison on speed}. We compare the average computational time (second) of different methods (except CC), which is shown in Table \ref{t2}. All experiments are run under the Matlab environment on a machine with 3.5GHz CPU and 32GB RAM. The fastest result is highlighted in bold. One can see that Neat Image (NI) is the fastest and costs about 0.9 second, while the proposed MC-WNNM needs 202.9 seconds. Noted that CBM3D, TNRD, and NC are implemented with compiled C++ mex-function and with parallelization, while WNNM, MLP, DnCNN, and the proposed MC-WNNM are implemented purely in Matlab.

\vspace{-1mm}
\section{Conclusion}
\vspace{-1mm}

The real noisy color images have different noise statistics across the R, G, B channels due to digital camera pipelines in CCD or CMOS sensors.\ This makes the real color image denoising problem more challenging than grayscale image denoising.\ In this paper, we proposed a novel multi-channel (MC) denoising model to effectively exploit the redundancy across color channels while differentiating their different noise statistics.\ Specifically, we introduced a weight matrix to the data term in the RGB channel concatenated weighted nuclear norm minimization (WNNM) model, and the resulting MC-WNNM model can process adaptively the different noise in RGB channels.\ We solved the MC-WNNM model via an ADMM algorithm.\ Extensive experiments on synthetic and real datasets demonstrated that the proposed MC-WNNM method outperforms significantly the other competing denoising methods.

The proposed MC-WNNM model can be extended in at least two directions.\ Firstly, it is worthy to investigate new weight matrix beyond the diagonal form, such as the correlation form \cite{nearcor}, to further improve the color image denoising performance.\ Secondly, the proposed MC-WNNM model can be extended for hyperspectral image analysis, which may contain hundreds of bands with complex noise statistics.

{
\huge
\bibliographystyle{unsrt}
\bibliography{egbib}
}

\end{document}